\documentclass[conference]{IEEEtran}
\IEEEoverridecommandlockouts
\usepackage{cite}
\usepackage{amsmath,amssymb,amsfonts}
\usepackage{algorithmic}
\usepackage{graphicx}
\usepackage{textcomp}
\usepackage{xcolor}
\usepackage{multirow}
\usepackage{booktabs}
\usepackage{pifont}

\usepackage{rotating}

\def\BibTeX{{\rm B\kern-.05em{\sc i\kern-.025em b}\kern-.08em
    T\kern-.1667em\lower.7ex\hbox{E}\kern-.125emX}}
\begin{document}

\title{Hierarchical Collaborative Fusion for 3D Instance-aware Referring Expression Segmentation}

\author{
    \IEEEauthorblockN{
        Keshen Zhou\IEEEauthorrefmark{2}\textsuperscript{,1}, 
        Runnan Chen\IEEEauthorrefmark{2}\textsuperscript{,2,*}\thanks{* Corresponding author.}, 
        Mingming Gong\IEEEauthorrefmark{3}\IEEEauthorrefmark{4}\textsuperscript{,3}, 
        Tongliang Liu\IEEEauthorrefmark{2}\IEEEauthorrefmark{4}\textsuperscript{,4,*}
    }
    \IEEEauthorblockA{
        \IEEEauthorrefmark{2}The University of Sydney\\
        \IEEEauthorrefmark{3}The University of Melbourne\\
        \IEEEauthorrefmark{4}Mohamed bin Zayed University of Artificial Intelligence
    }    
    \IEEEauthorblockA{
        \textsuperscript{1}kzho6770@uni.sydney.edu.au;
        \textsuperscript{2}crnsmile@connect.hku.hk;
        \textsuperscript{3}mingming.gong@unimelb.edu.au;\\
        \textsuperscript{4}tongliang.liu@sydney.edu.au
    }
}

\maketitle

\begin{abstract}
Generalised 3D Referring Expression Segmentation (3D-GRES) localizes objects in 3D scenes based on natural language, even when descriptions match multiple or zero targets. Existing methods rely solely on sparse point clouds, lacking rich visual semantics for fine-grained descriptions. We propose HCF-RES, a multi-modal framework with two key innovations. First, Hierarchical Visual Semantic Decomposition leverages SAM instance masks to guide CLIP encoding at dual granularities—pixel-level and instance-level features—preserving object boundaries during 2D-to-3D projection. Second, Progressive Multi-level Fusion integrates representations through intra-modal collaboration, cross-modal adaptive weighting between 2D semantic and 3D geometric features, and language-guided refinement. HCF-RES achieves state-of-the-art results on both ScanRefer and Multi3DRefer.
\end{abstract}


\section{Introduction}
\label{sec:intro} 

3D Referring Expression Segmentation (3D-RES) aims to segment target objects in 3D point cloud scenes based on natural language descriptions~\cite{old_refer:1}. 
Its generalized formulation, 3D-GRES~\cite{3dgres_compare:1,dataset:multi3drefer,base_refer:2}, further extends the task to handle expressions that may refer to multiple objects or no objects at all. 
The tasks require bridging the gap between linguistic semantics and 3D spatial-geometric representations to produce more accurate segmentation masks, essential for applications such as robotic manipulation, augmented reality, and embodied AI.

Recent methods have evolved from two-stage pipelines~\cite{old_refer:1} to end-to-end architectures~\cite{res:segpoint,res:unified3d,rg-san,base_refer:1,base_refer:2,base_refer:3,res:reason3d} that directly predict segmentation masks through query-based mechanisms. 
Among these, superpoint-based frameworks~\cite{base_refer:1,base_refer:2,rg-san,base_refer:3,res:reason3d} have become prevalent, leveraging geometric priors from pre-clustered point groups to improve both efficiency and accuracy. 

\begin{figure}
    \centering
    \resizebox{\columnwidth}{!}{
    \includegraphics[]{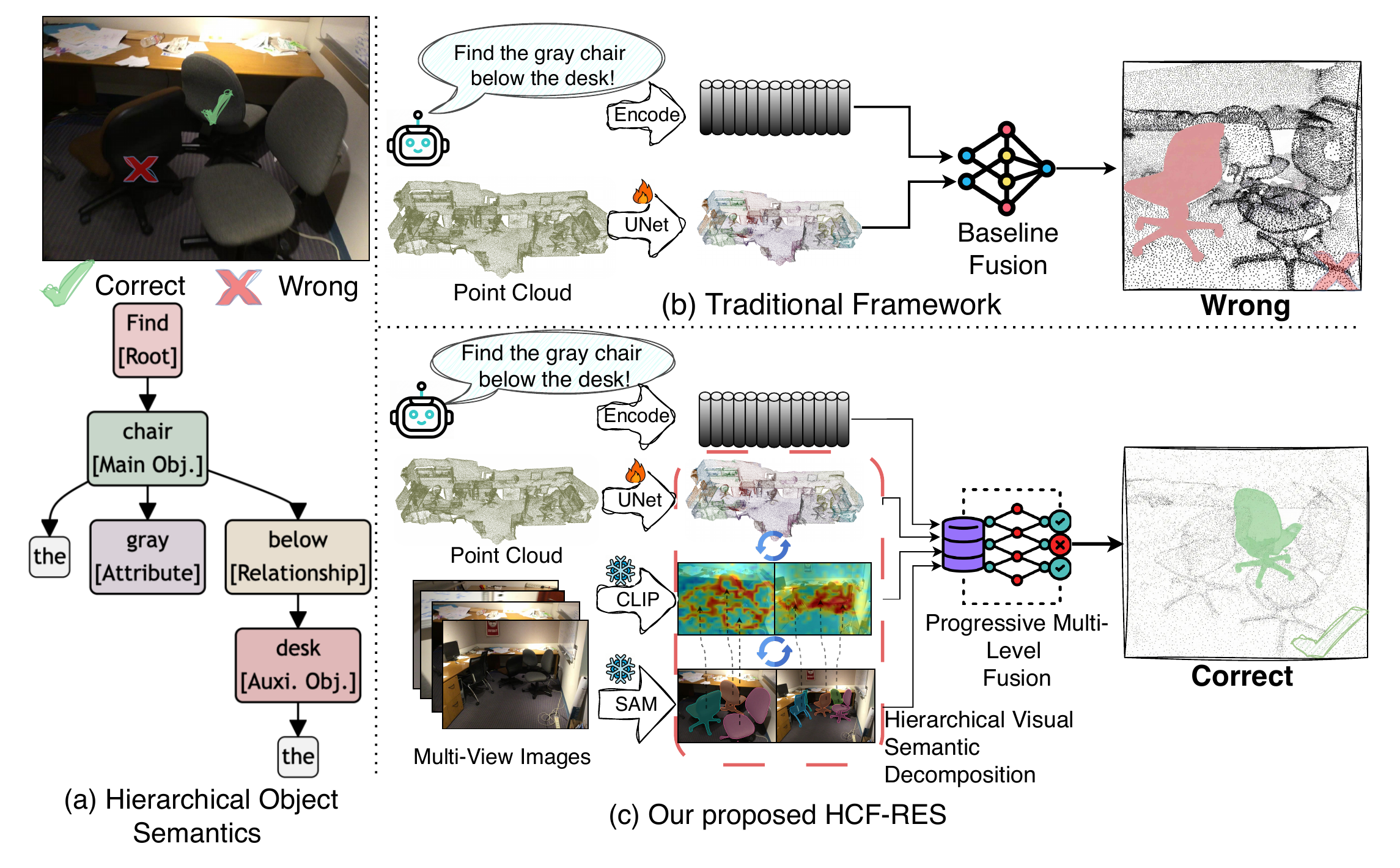}}
    \caption{
    (a) Language descriptions inherently encode hierarchical object semantics (e.g., main object "chair" with attributes "gray" and relationship "below desk").
    (b) Traditional methods rely solely on sparse point clouds, leading to wrong segmentation. 
    (c) Our method leverages pre-trained vision models (CLIP\cite{clip} and SAM\cite{Samv1}) on filtered multi-view images to extract rich semantic features, achieving better alignment across images, point clouds, and natural language.}
    \label{fig:teaser}
\end{figure}

Despite these advances, these methods predominantly rely on 3D point cloud data alone, which inherently suffers from the limitation of sparsity and lacks rich texture features crucial for interpreting complex visual attributes described in natural language (e.g. \textit{the gray chair} in Fig \ref{fig:teaser}(a)).
While point clouds captured by depth-aware sensors such as LiDAR encode precise geometry but remain sparsely distributed and largely color-blind, multi-view RGB images provide dense texture information essential for understanding fine-grained visual descriptions. 
This raises a critical question: How to integrate multi-view RGB images, LiDAR point clouds and natural languages for efficient 3D Referring Expression Segmentation?

Current multi-modal fusion approach~\cite{base_refer:3} has explored leveraging CLIP\cite{clip} to inject 2D image features into 3D scenes (see Fig~\ref{fig:teaser}(b)). 
However, this approach primarily fuses at the pixel level without leveraging the object-level semantics, treating all spatial regions equally rather than adopting \textit{collaborative fusion} strategies that distinguish between different semantic entities.
This limitation fails to distinguish multiple object instances within the same image (i.e. not \textit{instance-aware}).
When pixel-level features are projected and aggregated into 3D superpoints, features from different semantic entities inevitably intermingle, hindering cross-modal alignment, as language descriptions inherently encode hierarchical object-level semantics (Fig \ref{fig:teaser}(a)). 

To address these limitations, we propose a \textbf{H}ierarchical multi-modal \textbf{C}ollaborative \textbf{F}usion framework for 3D Instance-aware RES (HCF-RES in Fig \ref{fig:teaser}(c)).
First, we introduce \textbf{Hierarchical Visual Semantic Decomposition}, which leverages SAM~\cite{Samv1} to segment high-quality instance masks from multi-view images and employs CLIP~\cite{clip} encoder to extract visual-semantic features at two complementary granularities: dense pixel-level features from entire images and instance-level features through masked pooling guided by SAM segmentation, preserving object boundaries throughout 2D-to-3D projection.

Second, we propose \textbf{Progressive Multi-level Fusion}, a three-stage strategy that achieves collaborative cross-modal alignment: 
(1) intra-modal integration establishes complementarity between pixel-level and instance-level features within the 2D visual domain; 
(2) cross-modal integration learns adaptive weighting between 2D semantic and 3D geometric features at each superpoint, dynamically determining the contribution of each modality;
(3) \textbf{Language-guided instance refinement} enhances selected queries with instance-aware semantics for precise alignment with linguistic descriptions.

In summary, our main contributions are as follows:
\begin{itemize}
    \item We propose {Hierarchical Visual Semantic Decomposition}, which leverages SAM-segmented instance masks to guide CLIP encoding at two complementary granularities: dense pixel-level features from entire images and instance-level features through masked encoding.
    \item We introduce a progressive multi-level fusion strategy that achieves collaborative cross-modal alignment through three stages: intra-modal integration of visual features, cross-modal dynamic weighting and language-guided instance refinement.
    \item We demonstrate that HCF-RES achieves state-of-the-art performance on ScanRefer \cite{dataset:scanrefer} and Multi3DRefer \cite{dataset:multi3drefer} datasets.
\end{itemize}


\section{Related Work}
\subsection{3D Referring Expression Segmentation}
3D Referring Expression Segmentation extends 2D-RES~\cite{2dres:5} to localize and segment objects in 3D scenes based on natural language descriptions.
Early methods such as TGNN~\cite{old_refer:1} adopted a two-stage framework that first generates proposals and then matches them with textual descriptions. While producing reasonable results, such pipelines are computationally expensive and the decoupled modules limit end-to-end optimisation.

To address this, 3D-STMN~\cite{base_refer:1} introduced superpoints~\cite{superpoint,spformer} as geometric priors to align with textual features, establishing a framework adopted by subsequent works~\cite{base_refer:2,rg-san,base_refer:3,res:reason3d}.
3D-GRES~\cite{base_refer:2} further extended the task to generalized settings~\cite{3dgres_compare:1, dataset:multi3drefer} where expressions may refer to multiple objects or no objects.
However, existing approaches in 3D-RES predominantly rely on geometric features from point clouds, lacking sophisticated multi-modal features and fusion strategies that can effectively leverage rich semantic information from pre-trained vision-language models~\cite{clip,Samv1}.
Therefore, our focus is to explore multi-modal fusion strategies that incorporate both semantic-rich vision-language models and geometry-aware 3D features for robust referring expression segmentation.

\subsection{Multi-modal 3D Fusion}
Multi-modal 3D fusion has emerged as a key technique for 3D scene understanding by integrating complementary information from point clouds and RGB images~\cite{runnan1,mllm-for3d}.
In 3D referring expression segmentation, effectively integrating 3D geometry, 2D visual semantics, and text descriptions remains challenging.

IPDN~\cite{base_refer:3} explores this direction by projecting CLIP features to point clouds and aggregating them into superpoints through geometric correspondence.
While effective, this pixel-level strategy, like most existing fusion approaches, operates primarily at the scene level without distinguishing foreground objects from background regions~\cite{runnan1}.
Recent works have identified the importance of leveraging fine-grained instance-level information and multi-attribute interactions for effective object disambiguation~\cite{vg:fourways}.
In contrast, our work develops a progressive multi-level fusion framework that leverages SAM-guided dual-granularity features~\cite{Samv1,clip} to incorporate precise instance boundary knowledge, enabling dynamic weighting mechanisms that address the inherent limitations of scene-level fusion.

\section{Methodology}
\subsection{Pipeline Overview}
\label{sec:pipeline overview} 
We address Generalised Referring Expression Segmentation (GRES) in 3D indoor scenes~\cite{dataset:multi3drefer}, where referring expressions may correspond to multiple objects, a single object, or no objects. Given a point cloud scene $P \in \mathbb{R}^{N_{p} \times C}$ and a language utterance $\mathcal{U}$, the goal is to produce a binary segmentation mask $\mathcal{M} \in \{0,1\}^{N_{p}}$ identifying all points belonging to the referred object(s).

Our framework builds upon prior 3D-GRES pipelines~\cite{base_refer:2} that combine geometric, visual, and linguistic information while addressing a fundamental challenge: effectively aligning multi-view RGB semantics with sparse 3D geometry to achieve language-grounded scene understanding.
To address this, our hierarchical visual semantic decomposition (\ref{sec:embeddings}) leverages instance masks from SAM to guide CLIP encoding at two complementary granularities: dense pixel-level features from the entire images and instance-level features through masked encoding.
Our progressive multi-level fusion strategy (Section~\ref{sec:fusion}) then integrates these representations through intra-modal collaboration of 2D features, adaptive cross-modal weighting, and language-guided instance refinement for precise query initialization.

\begin{figure}[t]
  \centering
 \includegraphics[width=1\linewidth]{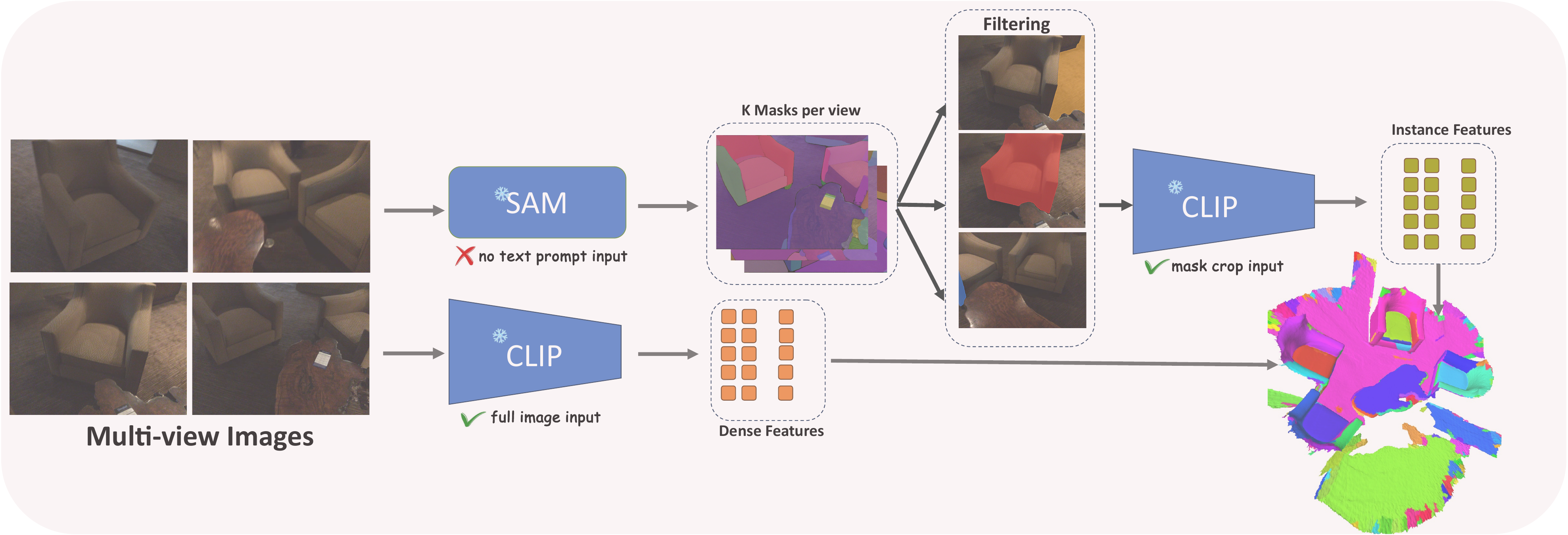}
   \vspace{-0.35cm}
   \caption{\textbf{Overview of our Hierarchical Visual Semantic Decomposition}: We employ the SAM \cite{Samv1} to segment the instances segmentation masks for each multi-view images without requiring annotations and each mask is then filtered by quality before encoding with CLIP \cite{clip} to obtain its instance-level and pixel-level features. These multi-granularity features are then subsequently projected to the point cloud and aggregated into superpoints representations.}
   \label{fig:multi-view}
\end{figure}

\begin{figure*}[t!]
  \centering
 \includegraphics[width=1\linewidth]{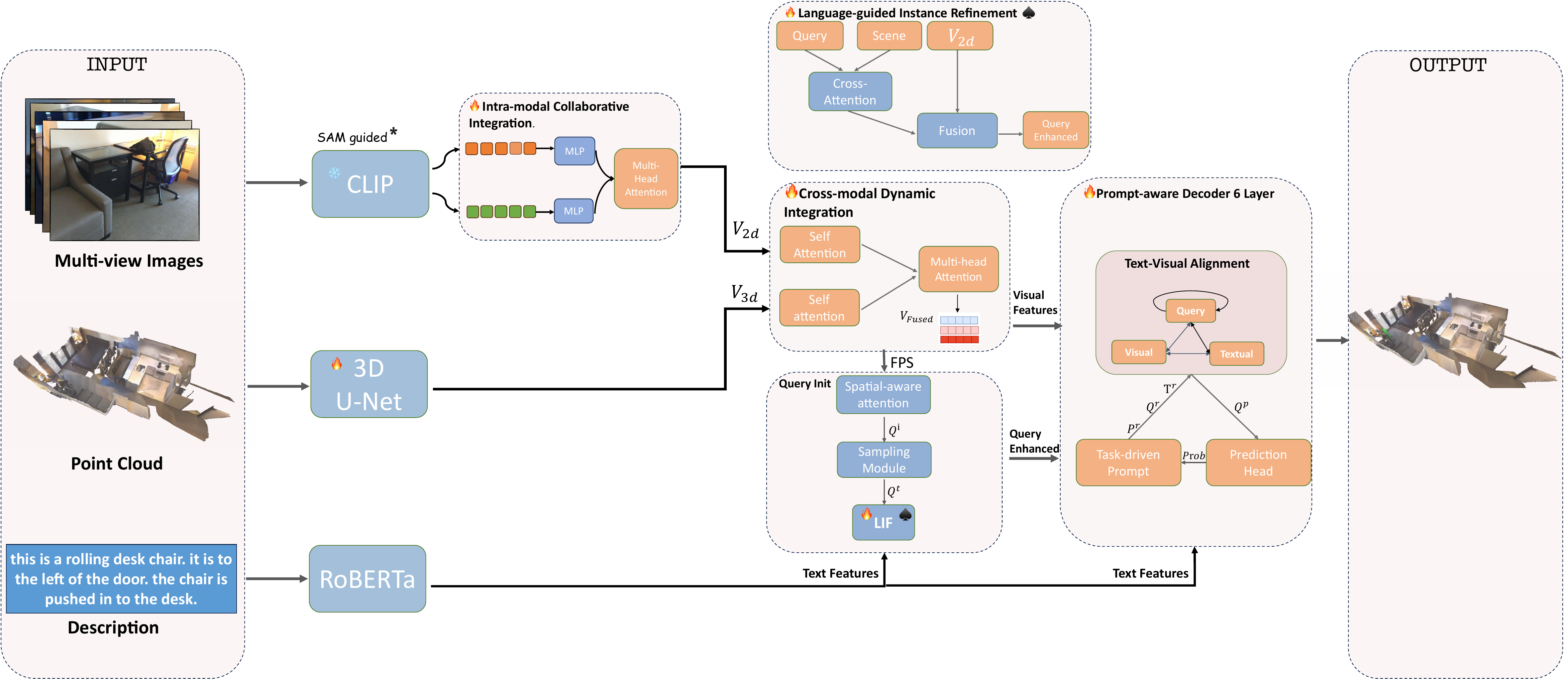}
   \caption{Pipeline of HCF-RES: The framework processes three input modalities: point clouds via 3D U-Net, text via RoBERTa, and multi-view images via CLIP guided by SAM \textsuperscript{*} through our Hierarchical Visual Semantic Decomposition (Section~\ref{SAM-based extraction} and Figure~\ref{fig:multi-view}). After cross-modal feature integration and language-guided sampling, our Language-guided Instance Refinement enhances selected queries through scene context awareness and 2D semantic fusion. Enhanced queries are decoded via a 6-layer decoder~\cite{base_refer:2} for final 3D referring expression segmentation.}
   \label{fig:pipeline}
\end{figure*}

\subsection{Preliminary}
\label{sec:prel}
\noindent{\bf Language Encoding.} 
We employ pre-trained RoBERTa \cite{roberta} to encode referring expressions into textual embeddings $T \in \mathbb{R}^{N_T \times D}$ where $N_T$ denotes the number of tokens and $D$ denotes the embedding dimension for cross-modal alignment. 
Following established practices~\cite{EDA}, we adopt scene graph parsing to decompose language expressions into structured hierarchical object semantics that enable fine-grained supervision of vision-language correspondence.

\noindent{\bf 3D Scene Representation.} 
We utilize a U-Net style backbone~\cite{backbone:3d} to extract point-wise features $\mathbf{F}_{p} \in \mathbb{R}^{N_{p} \times C}$, where $C$ denotes the channels. 
To reduce computational complexity, we feed these features into a superpoint pooling layer that leverage pre-computed superpoint~\cite{superpoint} to generate $N_S$ superpoints and aggregate point-wise features within each superpoint via average pooling to obtain $F_{3d} \in \mathbb{R}^{N_{S}\times C}$. 
To enable cross-modal alignment, we project the superpoint features into a shared embedding space using an adapter: 
\begin{equation}
V_{3d} = \text{Adapter}(F_{3d}),
\end{equation}
where $V_{3d} \in \mathbb{R}^{N_{S} \times D}$ represents the projected 3D visual features that serve as spatial anchors for subsequent multi-modal fusion. These features, along with textual embeddings, are used to initialize sparse queries for cross-modal alignment processing.

\subsection{Hierarchical Visual Semantic Decomposition}
\label{sec:embeddings}
\noindent{\bf Dense Multi-view Feature Encoding}
We leverage pre-trained CLIP models~\cite{clip} to extract dense visual-semantic features from multi-view RGB images $\{I_i\}_{i=1}^{N_I}$.
The CLIP encoder processes each image to extract patch-level features at intermediate layers, which are subsequently upsampled via bilinear interpolation to obtain dense pixel-level representations $\{F_i^{img} \in \mathbb{R}^{H \times W \times D}\}_{i=1}^{N_I}$.

These 2D features are projected into 3D space using camera parameters:
\begin{equation}
    \mathbf{p}_{\text{world}} = \mathbf{T}_{\text{cam}} \mathbf{K}^{-1} \begin{bmatrix} u \cdot d(u,v) \\ v \cdot d(u,v) \\ d(u,v) \end{bmatrix}.
\label{eq:projection}
\end{equation}
where $T_{cam}$is the camera extrinsic matrix, $\mathbf{K}$ is the intrinsic matrix, and $d(u,v)$ denotes the depth at pixel $(u, v)$.
The projected features are then aggregated into superpoint-level representations via spherical query and geometric pooling.

However, since CLIP is trained to encode entire images into unified representations via the global token $[CLS]$, these intermediate features lack 
explicit instance boundaries~\cite{openseg}. When features from different objects are projected and aggregated into the same superpoints, they inevitably intermingle, leading to ambiguous semantic representations. This motivates our SAM-guided instance-level feature extraction strategy.

\noindent{\bf Instance-aware semantic encoding.}
\label{SAM-based extraction}
To preserve object boundaries and prevent feature intermingling, we leverage 
SAM~\cite{Samv1} to segment high-quality instance masks from multi-view images. 
For each image $I_i$, SAM segments a set of instance masks $\mathcal{M} = \{M_1, M_2, \ldots, M_{n}\}\in \{0,1\}^{H \times W}$, accompanied by predicted IoU and stability scores, enabling quality-based filtering to retain only high-confidence instance segmentations. 
Since binary masks with hard boundaries would cause discontinuous feature weighting and lose information, we apply Gaussian blur to generate soft masks 
\begin{equation}
\tilde{M}_i = \mathcal{G}_\sigma * M_i
\end{equation}
where $\mathcal{G}_\sigma$ is a 2D Gaussian kernel with standard deviation $\sigma$, and $*$ denotes spatial convolution.
Rather than cropping individual instances which would discard valuable contextual information, we feed the complete RGB image through the CLIP visual encoder \cite{clip} to obtain spatial feature tokens $F_{spatial} \in \mathbb{R}^{N_{patch} \times D}$ that encode local semantic patterns while preserving full spatial context.
We resize the soft mask to the CLIP spatial token grid resolution via bilinear interpolation and compute instance-specific features through mask-weighted pooling:
\begin{equation}
f_{inst} = \frac{\sum_{j=1}^{N_{patch}} w_j \cdot f_j^{spatial}}{\sum_{j=1}^{N_{patch}} w_j}
\end{equation}
where $w_j$ represents the soft weight at token position $j$, effectively extracting semantic features from the instance region while maintaining smooth transitions at boundaries.
These instance-aware features $f_{inst} \in \mathbb{R}^{D}$ capture semantically coherent object representations while preserving contextual information from the complete image input.
When these instance-aware features are projected to 3D space following the same spherical querying and superpoint aggregation methods, each feature corresponds to a semantically coherent object region rather than arbitrary pixel patches, reducing inter-object semantic interference and preserving clear object boundaries in the final superpoint representations.
By combining both dense and instance-aware features, our multi-view semantic encoding provides complementary perspectives: dense features capture fine-grained local patterns while instance features preserve object-level semantic coherence, together forming a comprehensive representation for language grounding.

\begin{table}
    \centering
    \caption{3D-RES results on ScanRefer.}
    \resizebox{\linewidth}{!}
        {
        \begin{tabular}{l|c|ccc}
            \toprule 
            \multirow{2}{*}{Method} & \multirow{2}{*}{Venue} & \multicolumn{3}{c}{\textbf{Overall}} \\
            & & Acc@0.25 & Acc@0.5 & mIoU \\
            \midrule
            \midrule
            InstanceRefer  ~\cite{res:instancerefer} & ICCV2021 & 40.2 & 33.5 & 30.6  \\
            3D-STMN ~\cite{base_refer:1} & AAAI2024 & {54.6} & {39.8} & {39.5}  \\
            SegPoint ~\cite{res:segpoint} & ECCV2024 & - & - & {41.7} \\
            Reason3D ~\cite{res:reason3d} & 3DV2025 & {57.9} & {41.9} & {42.0} \\
            MCLN ~\cite{res:mcln} & ECCV2024 & {58.7} & {50.7} & {44.7}  \\
            RefMask3D ~\cite{res:he2024refmask3d} & ACMMM2024 & {55.9} & {49.2} & {44.9} \\
            MDIN ~\cite{base_refer:2} & ACMMM2024 & {58.0} & {53.1} & {48.3}  \\
            IPDN ~\cite{base_refer:3} & AAAI2025 & {59.9} & {54.4} & {49.5} \\
            \midrule
            HCF-RES & - & {\textbf{60.9}} & {\textbf{55.7}} & {\textbf{50.5}} \\  
            \bottomrule
        \end{tabular}
    }
    \label{tab:scanrefer_benchmark}
\end{table}

\subsection{Progressive Multi-level Fusion}
\label{sec:fusion}
Our fusion framework operates hierarchically: first aggregating multiple branches of 2D semantic features into a unified representation, then performing dynamic cross-modal fusion that adaptively balances the contributions from different modalities based on their local relevance.
This progressive fusion design ensures that both fine-grained instance semantics and spatial geometric structures are preserved throughout the integration process, ultimately producing a robust multi-modal representation for downstream query-based reasoning.

\noindent{\bf Intra-modal Collaborative Integration.} 
\label{2D-MLP}
Building upon the hierarchical visual semantic decomposition established above Section~\ref{sec:embeddings}, we address the challenge of effectively integrating complementary 2D representations through a collaborative fusion mechanism that leverages multi-head attention to dynamically balance dense spatial features and instance-aware semantics.
We process the dense pixel-level features $\mathbf{F}_{\text{dense}}$ and instance-aware features $\mathbf{F}_{\text{inst}}$ as separate branches at the superpoint level, where each branch preserves its distinctive semantic characteristics through dedicated neural pathways before collaborative integration.
Rather than naive concatenation that treats all semantic components equally, we decompose these complementary features into independent processing streams that maintain their semantic integrity while enabling dynamic interaction.
These processed features are integrated through multi-head attention that learns to dynamically weight different semantic aspects based on local context, enabling the model to emphasize dense spatial details versus instance-level coherence depending on the specific requirements of the referring expression:
\begin{equation}
V_{2d}^{fused} = \text{MultiHeadAttn}(\text{MLP}_{\text{dense}}(\mathbf{F}_{\text{dense}}), \text{MLP}_{\text{inst}}(\mathbf{F}_{\text{inst}}))
\end{equation}
This collaborative integration produces a unified 2D representation that preserves both fine-grained spatial details and instance-level semantic coherence, establishing a robust foundation for subsequent cross-modal dynamic integration with the 3D geometric features $V_{3d}$.

\begin{table*}[t!]
    \caption{Comparison of the 3D-GRES methods on Multi3DRefer. ZT, ST, and MT represent zero target, single target, and multiple targets, respectively. The left and right sides of the ``/" represent the situations with and without distractor objects, respectively.}
    \centering
    \setlength{\tabcolsep}{1mm}{
        \begin{tabular}{l|cccc|cccc|c}
            \toprule 
            & \multicolumn{4}{c|}{Acc@0.25} 
            & \multicolumn{4}{c|}{Acc@0.5} & \\
            \multirow{-2}{*}{Method}
            & ZT & ST & MT & All
            & ZT & ST & MT & All
            & \multirow{-2}{*}{mIoU}  \\
            \midrule

            ReLA~{\footnotesize\cite{3dgres_compare:1}} 
            & {36.2} / {72.7} & {48.3} / {83.4} & {73.0} & {61.8}
            & {36.2} / {72.7} & {20.4} / {65.5} & {42.4} & {37.4}
            & {42.8} \\

            M3DRef-CLIP~{\footnotesize \cite{dataset:multi3drefer}} 
            & {39.2} / {81.6} & {50.8} / {77.5} & {66.8} & {55.7}
            & {39.2} / {81.6} & {29.4} / {67.4} & {41.0} & {37.5}
            & {37.4} \\

            3D-STMN~{\footnotesize\cite{base_refer:1}} 
            & {42.6} / {76.2} & {49.0} / {77.8} & {68.8} & {60.4}
            & {42.6} / {76.2} & {24.6} / {69.2} & {43.9} & {40.9}
            & {43.0} \\
            
            MDIN~{\footnotesize\cite{base_refer:2}}
            & {22.2} / {67.2} & {54.2} / {86.6} & {76.5} & {65.4}
            & {22.2} / {67.2} & {27.0} / {71.6} & {45.6} & {41.8}
            & {45.8} \\ 

            IPDN~{\footnotesize\cite{base_refer:3}}
            & {36.8} / {81.1} & {59.9}  
            / {86.8} & {77.9} & {69.7} 
            & {36.8} / {81.1} & {35.3} 
            / {78.9} & {51.1} & {49.7}
            & {50.8} \\

            \hline
            \midrule
            
            HCF-RES
            & {\textbf{47.9}} / {\textbf{86.0}} & \textbf{{62.9}}  
            / \textbf{{88.4}} & \textbf{{78.9}} & \textbf{{72.3} }
            & \textbf{{47.9} }/ \textbf{{86.0}} & \textbf{{39.5} }
            / \textbf{{82.2}} & \textbf{{52.9}} & \textbf{{53.4}}
            & \textbf{{53.5}} \\ 
          
            \bottomrule
        \end{tabular}
    }
    \label{tab:multi3drefer}
\end{table*}

\noindent{\bf Cross-modal Dynamic Integration.} 
\label{2D3D-fusion}
While element-wise addition provides a baseline strategy for combining the unified 2D representation $V_{2d}^{fused}$ with 3D geometric features $V_{3d}$\cite{base_refer:3}, this approach fails to account for the varying reliability and contextual relevance of different modalities across spatial locations~\cite{vg:fourways}. 
To address this limitation, we introduce a spatially-adaptive weighting mechanism that learns to dynamically balance modal contributions by jointly examining both geometric and semantic characteristics at each superpoint location.
Specifically, the fusion module processes the concatenated features $[V_{2d}^{fused}; V_{3d}]$ through a lightweight neural network that predicts optimal blending weights for each modality:
\begin{equation}
V_{\text{unified}} = w_{2D} \odot V_{2d}^{fused} + w_{3D} \odot V_{3d}
\end{equation}
This spatially-adaptive weighting enables the model to emphasize 3D geometric features in regions where spatial relationships and geometric constraints are crucial, while prioritizing 2D semantic features in areas rich with visual attributes such as color, texture, and appearance details, where the blending weights $w_{2D}$ and $w_{3D}$ are learned parameters that adapt to local geometric and semantic characteristics, effectively leveraging the complementary strengths of both modalities to produce robust multi-modal representations for subsequent language-guided processing.

\noindent{\bf Language-guided Instance Refinement.} 
\label{instance-fusion}
While the unified multi-modal representation $V_{\text{unified}}$ provides a comprehensive foundation for cross-modal reasoning, performing complex instance-aware interactions directly across all superpoints would impose prohibitive computational overhead during training and inference. 
To address this efficiency challenge while maintaining alignment precision, we employ a progressive refinement strategy that first reduces the spatial complexity through farthest point sampling (FPS) \cite{fps} on 3d coordinates to select $N_{\text{seeds}}$ spatially diverse candidates, followed by language-guided selection \cite{base_refer:2,base_refer:3} that identifies the most relevant queries compared with text features encoded from above $T \in \mathbb{R}^{N_T \times D}$ for detailed instance-aware processing.
Building upon the sparse query initialization framework established in our scene representation, we apply cross-attention between candidate features and textual embeddings to compute language-visual relevance scores, selecting the top-k most semantically aligned queries $Q_{\text{selected}} \in \mathbb{R}^{m \times D}$ where $m \ll N_{\text{seeds}}$ for subsequent instance-guided enhancement.
At this computationally efficient scale, we introduce instance-guided enhancement that leverages the 2D instance features to enrich the selected queries through cross-attention mechanisms, enabling each query to discover semantic associations with scene-wide instance information while maintaining computational tractability.
This progressive refinement approach effectively balances computational efficiency with semantic precision, producing instance-aware queries that benefit from both global scene understanding and fine-grained object-level semantics for subsequent cross-modal reasoning in the decoder framework.

\subsection{Optimisation}
\label{sec:loss}
We follow an existing optimisation strategy~\cite{base_refer:2} for training. The training objectives include instance segmentation loss combining BCE and Dice terms, confidence estimation supervised by IoU scores, and vision-language alignment through contrastive learning.

\section{Experiments}

\subsection{Experimental Setting}

\noindent\textbf{Datasets.} We evaluate on ScanRefer~\cite{dataset:scanrefer} (51,583 expressions, 800 scenes) and Multi3DRefer~\cite{dataset:multi3drefer} (61,926 expressions including zero-target and multi-target cases). Metrics include mIoU and Acc@$k$IoU where $k \in \{0.25, 0.5\}$.

\noindent\textbf{Implementation.} We use Sparse 3D U-Net~\cite{backbone:3d} for point cloud encoding, RoBERTa~\cite{roberta} for language encoding, and frozen SAM~\cite{Samv1} and CLIP~\cite{clip} for 2D feature extraction. We use AdamW optimizer with initial learning rate $0.0001$ (decay power $4.0$) and batch size 16. The decoder consists of 6 layers with hidden dimension $D=256$. All experiments are conducted on one NVIDIA RTX 4090 GPU, introducing minimal additional overhead (see Table~\ref{tab:cost}).

\begin{figure}
  \label{fig:Qualitative}
  \centerline{\includegraphics[width=0.5\textwidth]{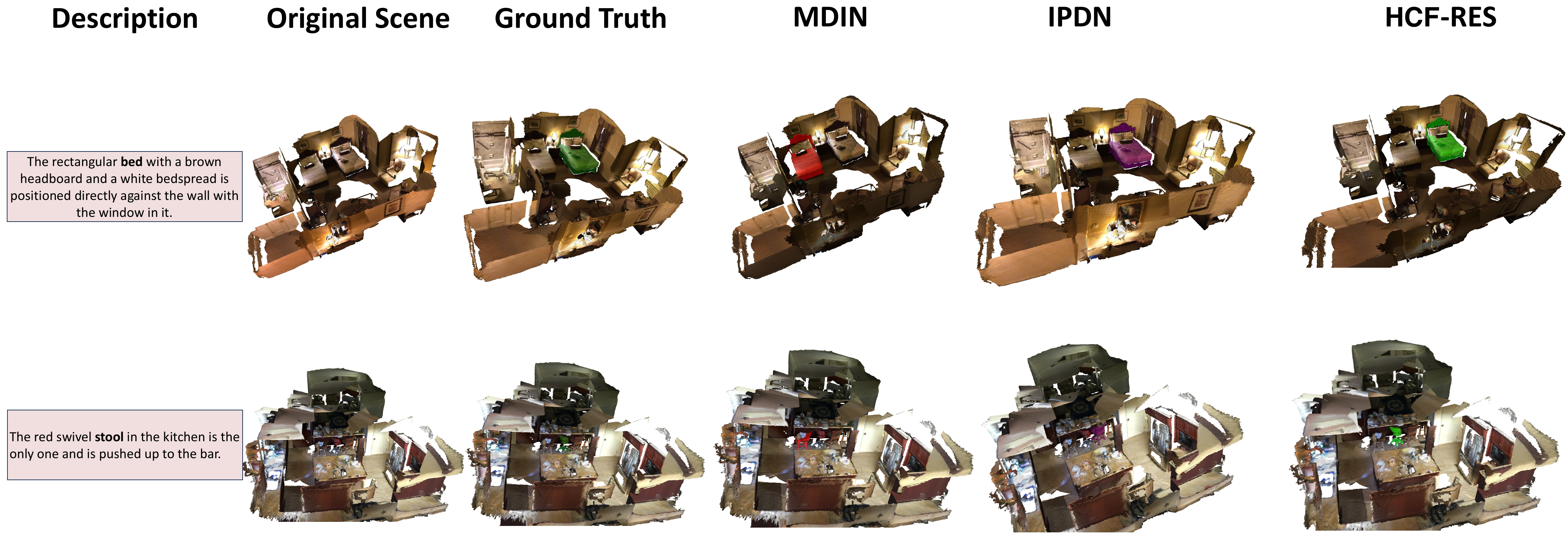}}
  \caption{Qualitative Result to compare our method HCF-RES with the state-of-art MDIN \cite{base_refer:2} and IPDN\cite{base_refer:3}. 
  Overall, our method produces more accurate segmentations. While IPDN achieves comparable results in general, it often shows noisy predictions near object boundaries.}
\end{figure}

\newcommand{\cmark}{\textcolor{green}{\ding{51}}} 
\newcommand{\xmark}{\textcolor{red}{\ding{55}}} 
\begin{table}
    \centering
    \caption{Ablation study on the proposed methods, evaluated on 3D-GRES. }
        \begin{tabular}{cc|ccc}
            \toprule
             \textbf{VSD} & \textbf{MLF}
            & \textbf{Acc@0.25} & \textbf{Acc@0.5} & \textbf{mIoU} \\
            \midrule
            \xmark &  \xmark
            & {70.9} & {50.5} & {51.5}  \\ 

            \xmark & \cmark 
            & {71.5} & {51.5} & {52.3} \\

            \cmark & \xmark  
            & {72.0} & {52.0} & {52.9} \\

            \cmark & \cmark  
            & {72.3}  & {53.4} & {53.5}\\ 
            \bottomrule
        \end{tabular}
    \label{compont_ablation}
\end{table}


\begin{table}[h!]
\centering
\caption{Inference speed and computational cost.}
\label{tab:cost}
    \resizebox{0.48\textwidth}{!}{
    \large
    \begin{tabular}{l c c c}
        \toprule
        Methods & Params (M) & Inference (ms)$\downarrow$ & mIoU$\uparrow$ \\
        \midrule
        IPDN~\cite{base_refer:3} & \textbf{156.53} & 535 & 50.8 \\
        \textbf{HCF-RES (ours)} & {157.69} & \textbf{523} & \textbf{53.5} \\
        \midrule
        $\Delta$ (ours $-$ IPDN) & +1.16 (+0.7\%) & $-$12 ($-$2.2\%) & +2.7 \\
        \bottomrule
    \end{tabular}
    }
    \vspace{-1em}
\end{table}

\subsection{Overall Performance}
To demonstrate the effectiveness of our model in the 3D referring segmentation task, we compare HCF-RES against state-of-the-art methods on the Multi3DRefer \cite{dataset:multi3drefer}  and ScanRefer\cite{dataset:scanrefer} validation set, as shown in Table ~\ref{tab:scanrefer_benchmark} and Table~\ref{tab:multi3drefer}.   
Our HCF-RES achieves the highest overall performance on Multi3DRefer \cite{dataset:multi3drefer} with 53.5 mIoU, surpassing the previous best method IPDN~\cite{base_refer:3} by 2.7 points and substantially outperforming MDIN~\cite{base_refer:2} by 7.7 points. 
Particularly noteworthy is our model's strong performance in zero target scenarios, achieving 47.9/86.0 Acc@0.25 with/without distractors compared to IPDN's 36.8/81.1 and MDIN's 22.2/67.2, demonstrating superior capability in distinguishing when no valid targets exist in the scene. 
HCF-RES also demonstrates strong performance in challenging multiple target scenarios, achieving 78.9 Acc@0.25 and 52.9 Acc@0.5, both representing the highest scores among all compared methods and highlighting our progressive fusion strategy's effectiveness in handling complex multi-object referring expressions.
On ScanRefer~\cite{dataset:scanrefer}, HCF-RES achieves 60.9\%/55.7\%/50.5\% on Acc@0.25/Acc@0.5/mIoU, establishing the best overall performance among compared methods.

\subsection{Ablation Study}
We conduct ablation studies on our proposed two modules, hierarchical visual semantic decompositions and progressive multi-level fusion on Multi3DRefer\cite{dataset:multi3drefer} validation set. 
As shown in Table \ref{compont_ablation}, both modules contribute to modest improvement. 
When comparing our different feature configurations, we observe that the hierarchical visual semantic decomposition (VSD)~\ref{sec:embeddings} provides more substantial gains than the progressive multi-level fusion (MLF) module~\ref{sec:fusion}  when applied individually.
Specifically, VSD alone achieves 72.0\% Acc@0.25 compared to 71.5\% with MLF alone, representing a 1.1\% versus 0.6\% improvement over the baseline respectively. The combination of both modules yields the optimal performance at 72.3\% Acc@0.25, 53.4\% Acc@0.5, and 53.5\% mIoU, demonstrating that both strategies work synergistically to enhance 3D RES accuracy. 

\section{Conclusion}
In this work, we present HCF-RES, a \emph{Hierarchical Multi-Modal Collaborative Fusion} framework for 3D Referring Expression Segmentation that addresses the hierarchical mismatch between instance-level linguistic semantics and pixel-level visual representations.
Our Hierarchical Visual Semantic Decomposition leverages SAM-segmented instance masks to guide CLIP encoding at dual granularities, extracting both dense pixel-level and instance-level features that preserve object boundaries during 2D-to-3D projection.
Our Progressive Multi-level Fusion integrates these representations through three stages: intra-modal collaboration, adaptive cross-modal weighting, and language-guided instance refinement, enabling effective alignment between 2D visual semantics and 3D geometric features.
%
Experiments on ScanRefer and Multi3DRefer show that HCF-RES achieves state-of-the-art results on both standard and \emph{generalised} (multi-/zero-target) splits, with particularly notable improvements in the zero-target scenarios, while introducing minimal computational overhead.

\bibliographystyle{IEEEbib}
\bibliography{icme2026references}

\vspace{12pt}

\clearpage

\end{document}